\documentclass[11pt,a4paper]{article}
\usepackage[hyperref]{emnlp-ijcnlp-2019}
\usepackage{times}
\usepackage{latexsym}
\usepackage{amsmath}
\usepackage{graphicx}
\usepackage{booktabs}
\usepackage[shortlabels]{enumitem}
\usepackage{url}
\usepackage{algorithm}
\usepackage{algorithmic}  
\usepackage{multirow}
\aclfinalcopy 

\title{Syntax-aware Multilingual Semantic Role Labeling}

\author{Shexia He$^{1,2,3,}$\thanks{$\ $ These authors made equal contribution.$^{\dag}$ Corresponding author. This paper was partially supported by National Key Research and Development Program of China (No. 2017YFB0304100) and Key Projects of National Natural Science Foundation of China (No. U1836222 and No. 61733011).}, Zuchao Li$^{1,2,3,*}$, Hai Zhao$^{1,2,3,\dag}$  \\
	$^{1}$Department of Computer Science and Engineering, Shanghai Jiao Tong University \\
	$^{2}$Key Laboratory of Shanghai Education Commission for Intelligent Interaction \\ and Cognitive Engineering, Shanghai Jiao Tong University, Shanghai, China\\
	$^{3}$MoE Key Lab of Artificial Intelligence, AI Institute, Shanghai Jiao Tong University \\
	{\tt \{heshexia, charlee\}@sjtu.edu.cn, zhaohai@cs.sjtu.edu.cn}
}

\date{}

\begin{document}
\maketitle
\begin{abstract}
  Recently, semantic role labeling (SRL) has earned a series of success with even higher performance improvements, which can be mainly attributed to syntactic integration and enhanced word representation. However, most of these efforts focus on English, while SRL on multiple languages more than English has received relatively little attention so that is kept underdevelopment. Thus this paper intends to fill the gap on multilingual SRL with special focus on the impact of syntax and contextualized word representation. Unlike existing work, we propose a novel method guided by syntactic rule to prune arguments, which enables us to integrate syntax into multilingual SRL model simply and effectively. We present a unified SRL model designed for multiple languages together with the proposed uniform syntax enhancement. Our model achieves new state-of-the-art results on the CoNLL-2009 benchmarks of all seven languages. Besides, we pose a discussion on the syntactic role among different languages and verify the effectiveness of deep enhanced representation for multilingual SRL.
\end{abstract}

\section{Introduction}

Semantic role labeling (SRL) aims to derive the meaning representation such as an instantiated predicate-argument structure for a sentence. The currently popular formalisms to represent the semantic predicate-argument structure are based on dependencies and spans. Their main difference is that dependency SRL annotates the syntactic head of argument rather than the entire constituent (span), and this paper will focus on the dependency-based SRL. Be it dependency or span, SRL plays a critical role in many natural language processing (NLP) tasks, including information extraction \cite{Christensen2011An}, machine translation \cite{xiong2012} and question answering \cite{yih-EtAl:2016}.

Almost all of traditional SRL methods relied heavily on syntactic features, which suffered the risk of erroneous syntactic input, leading to undesired error propagation. To alleviate this inconvenience, researchers as early as \citet{zhou-xu2015} propose neural SRL models without syntactic input. \citet{cai2018full} employ the biaffine attentional mechanism \cite{dozat2017deep} for dependency-based SRL. In the meantime, a series of studies \cite{roth2016,marcheggianiEMNLP2017, Strubell2018, li2018emnlp} have introduced syntactic clue in creative ways for further performance improvement, which achieve favorable results. However, applying the $k$-order syntactic tree pruning of \citet{he:2018Syntax} to the biaffine SRL model \cite{cai2018full} does not boost the performance as expected, which indicates that exploiting syntactic clue in state-of-the-art SRL models still deserves deep exploration.

Besides, most of SRL literature is dedicated to impressive performance gains on English and Chinese, but other multiple languages have received relatively little attention. We even observe that to date the best reported results of some languages (Catalan and Japanese) are still from the initial CoNLL-2009 shared task \cite{hajivc-EtAl2009}. Therefore, we launch this multilingual SRL study to fill the obvious gap ignored since a long time ago. Especially, we attempt to improve the overall performance of multilingual SRL by incorporating syntax and introducing contextualized word representation, and explore syntactic effect on other multiple languages.

Multilingual SRL needs to be carefully handled for the diversity of syntactic and semantic representations among quite different languages. Despite such a diversity, in this paper, we manage to develop a simple and effective neural SRL model to integrate syntactic information by applying argument pruning method in a uniform way. Specifically, we introduce new pruning rule based on syntactic parse tree unlike the $k$-order pruning of \citet{he:2018Syntax}, which is only simply determined by the relative distance of predicate and argument. Furthermore, we propose a novel method guided by syntactic rule to prune arguments for dependency SRL, different from the existing work. With the help of the proposed pruning method, our model can effectively alleviate the imbalanced distribution of arguments and non-arguments, achieving faster convergence during training. 

To verify the effectiveness and applicability of the proposed method, we evaluate the model on all seven languages of CoNLL-2009 datasets. Experimental results indicate that our argument pruning method is generally effective for multilingual SRL over our unified modeling. Moreover, our model using contextualized word representation achieves the new best results on all seven datasets, which is the first overall update since 2009. To the best of our knowledge, this is the first attempt to study seven languages comprehensively in deep learning models.

\begin{figure*}
	\centering
	\includegraphics[scale=0.9]{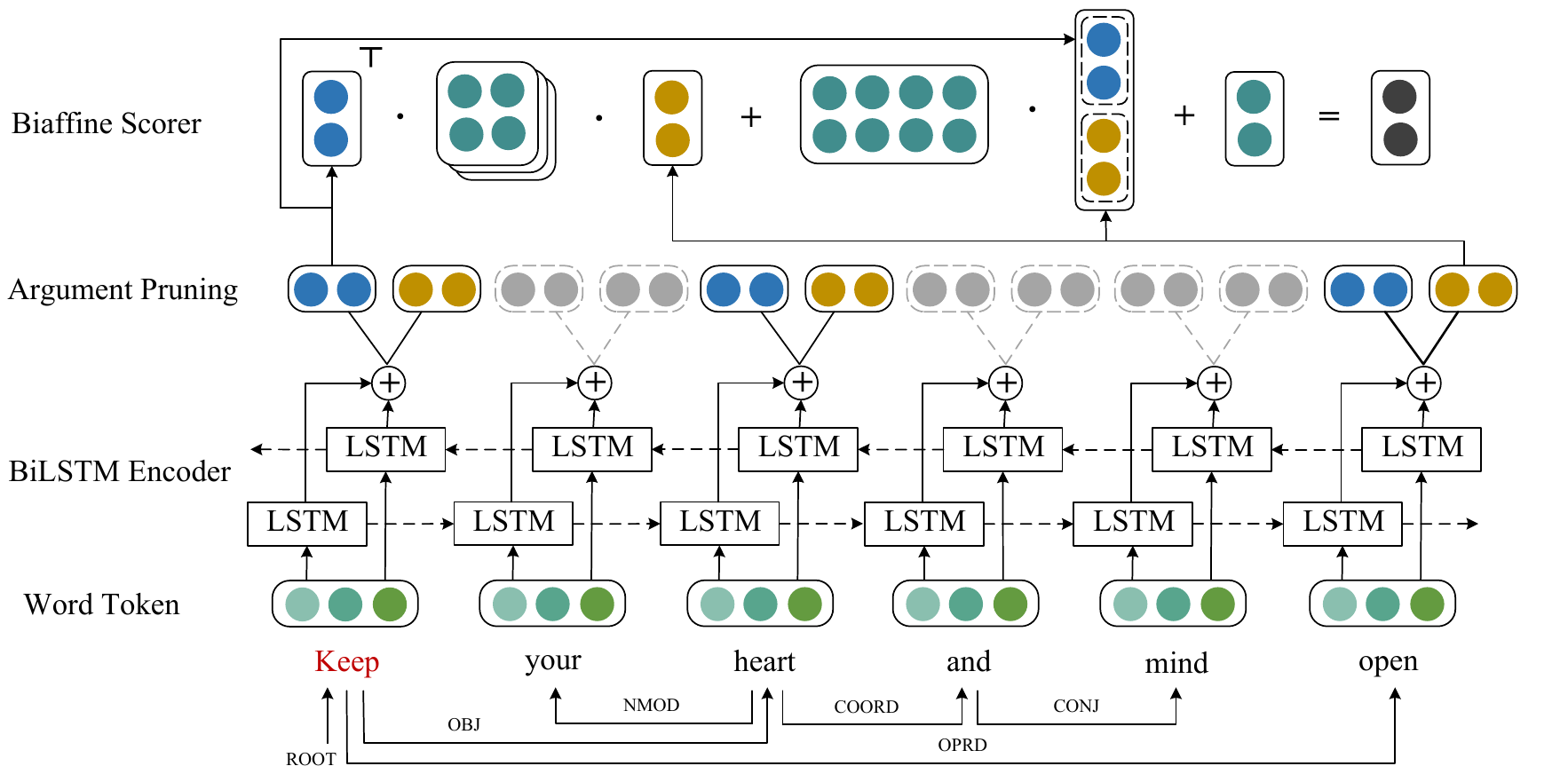}
	\caption{\label{fig:overview} Overall architecture of our SRL model. Red denotes the given predicate, and gray indicates that these units are dropped according to syntactic rule. The bottom is syntactic dependency.}
\end{figure*}

\section{Model}
Given a sentence, SRL can be decomposed into four classification subtasks, predicate identification and disambiguation, argument identification and classification. Since the CoNLL-2009 shared task has indicated all predicates beforehand, we focus on identifying arguments and labeling them with semantic roles. Our model builds on a recent syntax-agnostic SRL model \cite{cai2018full} by introducing argument pruning and enhanced word representation. In this work, we handle argument identification and classification in one shot, treating the SRL task as word predicate-argument pair classification. Figure \ref{fig:overview} illustrates the overall architecture of our model, which consists of three modules, (1) a bidirectional LSTM (BiLSTM) encoder, (2) an argument pruning layer which takes as input the BiLSTM representations, and (3) a biaffine scorer which takes as input the predicate and its argument candidates.

\subsection{BiLSTM Encoder}
Given a sentence and marked predicates, we adopt the bidirectional Long Short-term Memory neural network (BiLSTM) \cite{Hochreiter1997} to encode sentence, which takes as input the word representation. Following \citet{cai2018full}, the word representation is the concatenation of five vectors: randomly initialized word embedding, lemma embedding, part-of-speech (POS) tag embedding, pre-trained word embedding and predicate-specific indicator embedding.

Besides, the latest work \cite{li2018emnlp} has demonstrated that the contextualized representation ELMo (Embeddings from Language Models) \cite{ELMo} could boost performance of dependency SRL model on English and Chinese. To explore whether the deep enhanced representation can help other multiple languages, we further enhance the word representation by concatenating an external embedding from the recent successful language models, ELMo and BERT (Bidirectional Encoder Representations from Transformers) \cite{bert2018}, which are both contextualized representations. It is worth noting that we use ELMo or BERT to obtain pre-trained contextual embeddings rather than fine-tune the model, which are fixed contextual representations. 

\subsection{Argument Pruning Layer}
For word pair classification modeling, one major performance bottleneck is caused by unbalanced data, especially for SRL, where more than 90\% of argument candidates are non-arguments. A series of pruning methods are then proposed to alleviate the imbalanced distribution, such as the $k$-order pruning \cite{he:2018Syntax}. However, it does not extend well to other languages, and even hinders the syntax-agnostic SRL model as \citet{cai2018full} has experimented with different $k$ values on English. The reason might be that this pruning method breaks up the whole sentence, leading the BiLSTM encoder to take the incomplete sentence as input and fail to learn sentence representation sufficiently.

To alleviate such a drawback from the previous syntax-based pruning methods, we propose a novel pruning rule extraction method based on syntactic parse tree, which generally suits multilingual cases at the same time. In detailed model implementation, we add an argument pruning layer guided by syntactic rule following BiLSTM layers, which can absorb the syntactic clue simply and effectively. 

\paragraph{Syntactic Rule}
Considering that all arguments are predicate-specific instances, it has been generally observed that the distances between predicate and its arguments on syntactic tree are within a certain range for most languages. Therefore, we introduce language-specific rule based on syntactic dependency parses to prune some unlikely arguments, henceforth syntactic rule. Specifically, given a predicate $p$ and its argument $a$, we define $d_p$ and $d_a$ to be the distance from $p$ and $a$ to their nearest common ancestor node (namely, the root of the minimal subtree which includes $p$ and $a$) respectively. For example, $0$ denotes that predicate or argument itself is their nearest common ancestor, while $1$ represents that their nearest common ancestor is the parent of predicate or argument. Then we use the distance tuple ($d_p$, $d_a$) as their relative position representation inside the parse tree. Finally, we make a list of all tuples ordered according to how many times that each distance tuple occurs in the training data, which is counted for each language independently. 

It is worth noting that our syntactic rule is determined by the top-$k$ frequent distance tuples. During training and inference, the syntactic rule takes effect by excluding all candidate arguments whose predicate-argument relative position in parse tree is not in the list of top-$k$ frequent tuples.

\begin{figure}
	\centering
	\includegraphics[scale=1.2]{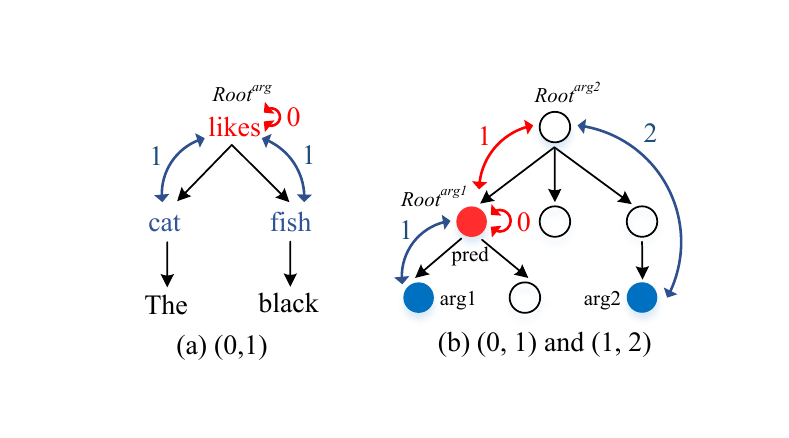}
	\caption{\label{fig:multi-syntax} Syntactic parse tree examples (dependency relations are omitted). Red represents the current predicate, and blue indicates its arguments.}
\end{figure}

Figure \ref{fig:multi-syntax} shows simplified examples of syntactic dependency tree. Given an English sentence in (a), the current predicate is \textit{likes}, whose arguments are \textit{cat} and \textit{fish}. For \textit{likes} and \textit{cat}, the predicate (\textit{likes}) is their common ancestor (denoted as $Root^{arg}$) according to the syntax tree. Therefore, the relative position representation of predicate and argument is $(0,1)$, so it is for \textit{likes} and \textit{fish}. As for the right one in (b), suppose the marked predicate has two arguments$-$$arg1$ and $arg2$, the common ancestors of predicate and arguments are respectively $Root^{arg1}$ and $Root^{arg2}$. In this case, the relative position representations are $(0,1)$ and $(1,2)$.  

\paragraph{Argument Pruning Method} 
To maintain the integrity of sequential inputs from the whole sentence, we propose a novel syntax-based method to prune arguments, unlike most existing work \cite{xue2004,Zhao2009Conll,he:2018Syntax} which prunes argument candidates in the pre-processing stage. As shown in Figure \ref{fig:overview}, the way to perform argument pruning strategy is very straightforward. In the argument pruning layer, our model drops these candidate arguments (more exactly, BiLSTM representations) which do not comply with the syntactic rule. In other words, only the predicates and arguments that satisfy the syntactic rule will be output to next layer.

For example (in Figure \ref{fig:overview}), given the sentence \textit{Keep your heart and mind open}, the predicate \textit{Keep} and the corresponding syntactic dependency (bottom), by definition, $(0,1)$ is inside the syntactic rule on this occasion. Therefore, these candidate arguments (i.e., \textit{your}, \textit{and}, \textit{mind}) will be pruned by the argument pruning layer. 

\subsection{Biaffine Scorer}
As mentioned above, our model treats SRL task as a word pair classification problem, tackling argument identification and classification in one shot. To label arguments of given predicates, we employ a scorer with biaffine attention \cite{dozat2017deep} (biaffine scorer for short) as role classifier on top of argument pruning layer for the final prediction, similar to \citet{cai2018full}. Biaffine scorer takes as input the BiLSTM hidden states of predicate and candidate arguments filtered by argument pruning layer, denoted by $h_p$ and $h_a$ respectively, and then computes the probability of corresponding semantic labels using biaffine transformation as follows:
\begin{align}
&\Phi_{r}(p, a) = \textit{Biaffine}(h_p, h_a) \nonumber\\
&= \{h_p\}^T \textbf{W}_1 h_a + \textbf{W}_2^T (h_p \oplus h_a) + \textbf{b} \nonumber
\end{align} 
where $\oplus$ represents concatenation operator, $\textbf{W}_1$ and $\textbf{W}_2$ denote the weight matrix of the bilinear and the linear terms respectively, and $\textbf{b}$ is the bias item. Note that the predicate itself is also included in its own argument candidate list and will be applied to compute scores, because a nominal predicate sometimes takes itself as its own argument.

\section{Experiments}

Our model\footnote{The code is available at \url{https://github.com/bcmi220/multilingual_srl}.} is evaluated on the CoNLL-2009 benchmark datasets, including Catalan, Chinese, Czech, English, German, Japanese and Spanish. The statistics of the training datasets can be seen in Table \ref{tab:statistics}. For the predicate disambiguation task, we follow previous work, using models \cite{Zhao2009Conll} for Catalan and Spanish, and the ones \cite{bjorkelund2009} for other languages. Besides, we use the officially predicted POS tags and syntactic parses provided by CoNLL-2009 shared-task for all languages.\footnote{There were two tracks in the CoNLL-2009 shared task, SRL-only and joint. For the former, all participants did not have to develop their own syntactic parsers and focused on the SRL model development, while for the latter, the participants had to build their own syntactic parser as well. For the sake of focusing the SRL work, in this work, we will take the official syntax provided by CoNLL-2009.} As for the contextualized representation, ELMo, we employ the multilingual version from \citet{che-EtAl:2018:K18-2}. For BERT, this work uses the BERT-Base, Multilingual Cased model \cite{bert2018}. For syntactic rule in argument pruning layer, to ensure more than 99\% coverage of true arguments in pruning output, we use the top-$120$ distance tuples on Japanese and top-$20$ on other multiple languages for a better trade-off between computation and coverage.

\begin{table}
	\centering
	\setlength{\tabcolsep}{4pt}
	\begin{tabular}{lrrrr}
		\toprule
		Dataset&\#sent &\#token &\#pred &\#arg\\
		\midrule
		Catalan & 13,200 & 390,302 & 37,431 & 84,367 \\
		Chinese & 22,277 & 609,060 & 102,813 & 231,869 \\
		Czech & 38,727 & 652,544 & 414,237 & 365,255 \\
		English & 39,279 & 958,167 & 179,014 & 393,699 \\
		German & 36,020 & 648,677 & 17,400 & 34,276 \\
		Japanese & 4,393 & 112,555 & 25,712 & 43,957 \\
		Spanish & 14,329 & 427,442 & 43,824 & 99,054 \\
		\bottomrule
	\end{tabular}
	\caption{Training data statistics of sentences, tokens, predicates and arguments. \# denotes numbers.}\label{tab:statistics}
\end{table}

\begin{table*}[!htb]
	\centering
	\setlength{\tabcolsep}{10pt}
	\begin{tabular}{lcccccc}
		\toprule  
		\multirow{2}{*}{Model}&\multicolumn{3}{c}{English}&\multicolumn{3}{c}{Chinese}\cr  
		\cmidrule(lr){2-4} \cmidrule(lr){5-7} &P&R&F$_1$&P&R&F$_1$ \cr  
		\midrule
		\citet{Zhao2009Conll} &$-$&$-$&86.2&80.4&75.2&77.7 \cr
		\citet{bjorkelund2009} &88.6&85.2&86.9&82.4&75.1&78.6 \cr
		\citet{Fitzgerald2015} &$-$&$-$&87.3&$-$&$-$&$-$ \cr
		\citet{roth2016} &90.0&85.5&87.7&83.2&75.9&79.4 \cr
		\citet{marcheggiani2017}  &88.7&86.8&87.7&83.4&79.1&81.2 \cr
		\citet{marcheggianiEMNLP2017} &89.1&86.8&88.0&84.6&80.4&82.5 \cr
		\citet{he:2018Syntax} \small{(with ELMo)} &89.7&89.3&89.5&84.2&81.5&82.8 \cr
		\citet{cai2018full} &89.9&89.2&89.6&84.7&84.0&84.3 \cr
		\citet{li2018emnlp} \small{(with ELMo)} &90.3&89.3&89.8&84.8&81.2&83.0 \cr
		\citet{lizc2019srl} \small{(with ELMo)} &89.6&91.2&90.4&$-$&$-$&$-$ \cr
		\midrule
		Our baseline &89.30&89.93&89.61&82.88&85.26&84.05 \cr
		+ AP &89.96&89.96&89.96&84.60&84.50&84.55 \cr
		+ BERT &89.80&91.20&90.50&85.76&86.50&86.13 \cr
		+ AP + ELMo &90.00&90.65&90.32&84.44&84.95&84.70 \cr
		+ AP + BERT &\textbf{90.41}&\textbf{91.32}&\textbf{90.86}&\textbf{86.15}&\textbf{86.70}&\textbf{86.42} \cr 
		\bottomrule
	\end{tabular}
	\caption{Precision, recall and semantic F$_1$-score on CoNLL-2009 English in-domain data and Chinese test set.}\label{tab:english}
\end{table*}

\subsection{Model Setup}
In our experiments, all real vectors are randomly initialized, including 100-dimensional word, lemma, POS tag embeddings and 16-dimensional predicate-specific indicator embedding \cite{he:2018Syntax}. The pre-trained word embedding is 100-dimensional GloVe vectors \cite{penningtonEMNLP2014} for English, 300-dimensional fastText vectors \cite{grave2018learning} trained on Common Crawl and Wikipedia for other languages, while the dimension of ELMo or BERT word embedding is 1024. Besides, we use 3 layers BiLSTM with 400-dimensional hidden states, applying dropout with an 80\% keep probability between time-steps and layers. For biaffine scorer, we employ two 300-dimensional affine transformations with the ReLU non-linear activation, also setting the dropout probability to 0.2. During training, we use the categorical cross-entropy as objective, with Adam optimizer \cite{adam2015} initial learning rate $2e^{-3}$. All models are trained for up to 500 epochs with batch size 64.

\begin{table*}
	\centering
	\setlength{\tabcolsep}{5pt}
	\begin{tabular}{lccccccc}
		\toprule
		Model & Catalan & Chinese & Czech & English & German & Japanese & Spanish \\
		\midrule
		CoNLL-2009 ST best system  & 80.3 & 78.6 & 85.4 & 85.6 & 79.7 & 78.2 & 80.5 \\
		\citet{Zhao2009Conll} & 80.3 & 77.7 & 85.2 & 86.2 & 76.0 & 78.2 & 80.5 \\
		\citet{roth2016} & $-$ & 79.4 & $-$ & 87.7 & 80.1 & $-$ & 80.2 \\
		\citet{marcheggiani2017} & $-$ & 81.2 & 86.0 & 87.7 & $-$ & $-$ & 80.3 \\
		\citet{kasai2019syntax} & $-$ & $-$ & $-$ & 90.2 & $-$ & $-$ & 83.0 \\
		\citet{lizc2019srl} & $-$ & $-$ & $-$ & 90.4 & $-$ & $-$ & $-$ \\
		The best previously published & 80.3 & 84.3 & 86.0 & 90.4 & 80.1 & 78.2 & 83.0 \\
		\midrule
		Our baseline & 84.07 & 84.05 & 88.35 & 89.61 & 78.36 & 83.08 & 83.47 \\
		+ AP & 84.35 & 84.55 & 88.76 & 89.96 & 78.54 & 83.12 & 83.70 \\
		+ BERT & 84.88 & 86.13 & 89.06 & 90.50 & 80.68 & 83.57 & 84.50 \\
		+ AP + ELMo & 84.35 & 84.70 & 89.52 & 90.32 & 78.65 & 83.43 & 83.82\\
		+ AP + BERT & \textbf{85.14} & \textbf{86.42} & \textbf{89.66} & \textbf{90.86} & \textbf{80.87} & \textbf{83.76} & \textbf{84.60}\\
		\bottomrule
	\end{tabular}
	\caption{Semantic F$_1$-score on CoNLL-2009 in-domain test set. The first row is the best result of CoNLL-2009 shared task \cite{hajivc-EtAl2009}. The previously best  published results of Catalan and Japanese is from \citet{Zhao2009Conll}, Chinese from \citet{cai2018full}, Czech from \citet{marcheggiani2017}, English from \citet{lizc2019srl}, German from \citet{roth2016} and Spanish from \citet{kasai2019syntax}.}
	\label{tab:main-results}
\end{table*}

\subsection{Results and Discussion}
In Table \ref{tab:english}, we compare our single model (\textbf{AP} is an acronym for argument pruning) against previous work on English in-domain data and Chinese test set. Our baseline is a modification to the model of \citet{cai2018full} which uniformly handled the predicate disambiguation. For English, our baseline gives slightly weaker performance than the work of \citet{lizc2019srl}, which used ELMo and employed a sophisticated span selection model for predicting predicates and arguments jointly. Our model with the proposed argument pruning layer (\textbf{+ AP}) brings absolute improvements of 0.35\% and 0.5\% F$_1$ on English and Chinese, respectively, which is on par with the best published scores. Moreover, we introduce deep enhanced representation based on the argument pruning. Our model utilizing BERT (\textbf{+ AP + BERT}) achieves the new best results on English and Chinese benchmarks.

\begin{table*}[!htb]
	\centering
	\setlength{\tabcolsep}{6.3pt}
	\begin{tabular}{lcccccccccc}
		\toprule  
		\multirow{2}{*}{Model}&\multicolumn{2}{c}{Catalan}& \multicolumn{2}{c}{Chinese}&\multicolumn{2}{c}{English}&\multicolumn{2}{c}{German}&\multicolumn{2}{c}{Spanish}\cr  
		\cmidrule(lr){2-3} \cmidrule(lr){4-5} \cmidrule(lr){6-7}\cmidrule(lr){8-9}\cmidrule(lr){10-11} &PD&F$_1$&PD&F$_1$&PD&F$_1$&PD&F$_1$&PD&F$_1$ \cr  
		\midrule
		Our baseline &87.50&84.07&94.92&84.05&95.59&89.61&81.45&78.36&86.53& 83.47 \cr
		\midrule
		\textit{Biaffine SRL} &89.10&84.70&95.60&84.56&95.04&89.60&81.64&78.45&87.44&83.85 \cr
		+ AP &89.52&84.90&95.60&84.76&95.38&89.88&81.65&78.50&87.56&83.92 \cr
		+ AP + BERT &90.08&\textbf{86.04}&96.17&\textbf{86.90}&96.37&\textbf{91.00}&82.36&\textbf{81.14}&88.27&\textbf{85.15} \cr
		\bottomrule
	\end{tabular}
	\caption{Results of full end-to-end model. PD denotes the accuracy of predicate disambiguation. + AP represents \textit{biaffine SRL+Argument Pruning Layer}, while the last row indicates \textit{biaffine SRL+Argument Pruning Layer+BERT}.}\label{tab:pd}
\end{table*}

Table \ref{tab:main-results} presents all test results on seven languages of CoNLL-2009 datasets. So far, the best previously reported results of Catalan, Japanese and Spanish are still from CoNLL-2009 shared task. Compared with previous methods, our baseline yields strong performance on all datasets except German. Especially for Catalan, Czech, Japanese and Spanish, our baseline performs better than existing methods with a large margin of 3.5\% F$_1$ on average. Nevertheless, applying our argument pruning to the strong syntax-agnostic baseline can still boost the model performance, which demonstrates the effectiveness of proposed method. On the other hand, it indicates that syntax is generally beneficial to multiple languages, and can enhance the multilingual SRL performance with effective syntactic integration.

Besides, we report the scores of leveraging ELMo and BERT for multiple languages (the last three rows in Table \ref{tab:main-results}). The use of contextualized word representation further improves model performance, which overwhelmingly outperforms previously published best results and achieves the new state of the art in multilingual SRL for the first time. Furthermore, we find that ELMo promotes the overall performance of SRL model, but BERT gives more significant performance increase than ELMo on all languages, which suggests that BERT is better at enriching contextual information. More interestingly, we observe that the performance gains from the proposed argument pruning method or these deep enhanced representations are relatively marginal on Japanese, one possible reason is the relatively small size of its training set. This observation also indicates that ELMo or BERT is more suitable for learning on large annotated corpus.

\subsection{End-to-end SRL}
As mentioned above, we combine the predicate sense output of previous work to make results directly comparable, since the official evaluation script includes such prediction in the F$_1$-score calculation. However, predicate disambiguation is considered a simpler task with higher semantic F$_1$-score and deserves more further research. To this end, we present a full end-to-end neural model for multilingual SRL, namely \textit{Biaffine SRL}, following \citet{cai2018full}.

Unlike most of SRL work treating the predicate sense disambiguation and semantic role assignment tasks as independent, we jointly handle predicate disambiguation and argument labeling in one shot by introducing a virtual node \textit{$<$VR$>$} as the nominal semantic head of predicate. It should be noted that the predicate sense annotation of Czech and Japanese is simply the lemmatized token of the predicate, a one-to-one predicate-sense mapping. Therefore, we ignore them and conduct experiments on other five languages.

Table \ref{tab:pd} shows the results of end-to-end setting. Compared to the baseline, our full end-to-end model (\textit{Biaffine SRL}) yields slightly higher precision of predicate disambiguation as a whole, which gives rise to a corresponding gain of semantic F$_1$. What is more, our model (using argument pruning and BERT) reaches the highest scores on the five benchmarks. Besides, experiments indicate that argument pruning promotes role labeling performance while BERT significantly improves the performance of predicate disambiguation.

\section{Analysis}
In this section, we perform further analysis to better understand our model, exploring the impact of language features, syntactic rule and syntactic contribution for multilingual SRL. Since recent work well studied dependency SRL on English and Chinese, we focus on other five languages, and these analyses are performed on CoNLL-2009 test sets without using ELMo or BERT embeddings. 

\subsection{Effectiveness of Language Feature}
As \citet{marcheggiani2017} point out, POS tag information is highly beneficial for English. Consequently, we conduct an ablation study on in-domain test set to explore how the language features impact our model. Table \ref{tab:pos} reports the F$_1$ scores of model which removes POS tag or lemma from the baseline. Results show that omitting POS tag or lemma leads to slight performance degradation ($-0.5\%$ and $-0.32\%$ F$_1$ on average, respectively), indicating that both can help improve performance of multilingual SRL. Interestingly, we see a drop of 1.0\% F$_1$ for Japanese not using POS tag, which demonstrates its importance.

\begin{table}
	\centering
	\setlength{\tabcolsep}{3pt}
	\begin{tabular}{lccc}
		\toprule
		Dataset & baseline & w/o POS tag & w/o lemma\\
		\midrule
		Catalan & 84.07 & 83.83 \small{(-0.24)} & 83.60 \small{(-0.47)} \\
		Czech & 88.35 & 88.10 \small{(-0.25)} & 88.20 \small{(-0.15)} \\
		German & 78.36 & 77.80 \small{(-0.56)} & 78.12 \small{(-0.24)} \\
		Japanese & 83.08 & 82.02 \small{(-1.06)} & 82.80 \small{(-0.28)} \\
		Spanish & 83.47 & 83.15 \small{(-0.32)} & 83.00 \small{(-0.47)} \\
		\bottomrule
	\end{tabular}
	\caption{Ablation of POS tag and lemma on test set.}\label{tab:pos}
\end{table}

\begin{figure}
	\centering
	\includegraphics[scale=0.50]{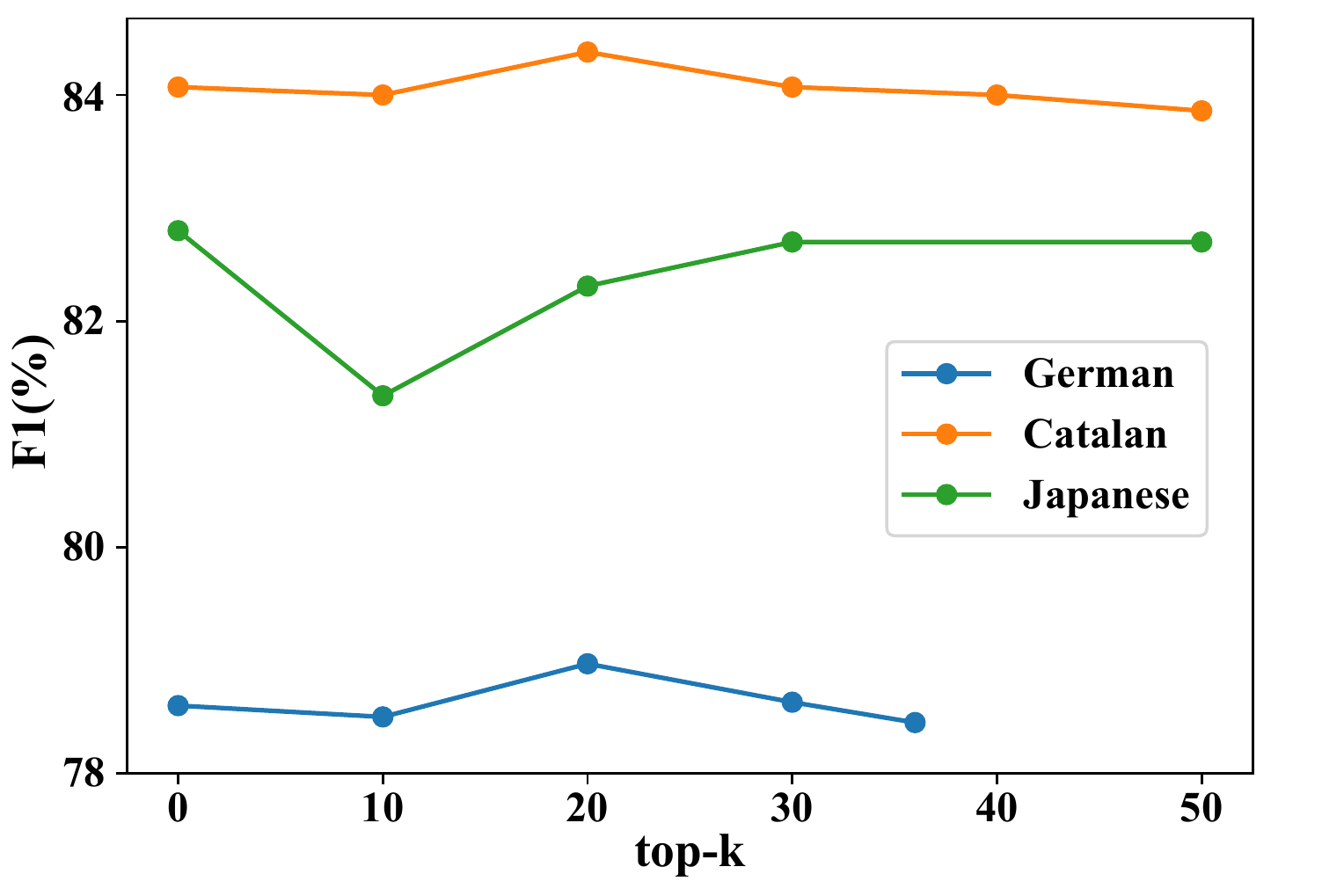}
	\caption{\label{fig:topk} F$_1$ scores on test set by top-$k$ argument pruning for German, Catalan and Japanese. }
\end{figure}

\subsection{Effectiveness of Syntactic Rule}
According to our statistics of syntactic rule on training data for seven languages, which is based on the automatically predicted parse provided by CoNLL-2009 shared task, the total number of distance tuples in syntactic rule is no more than 120 in these languages except that Japanese is about 260. It is in favor of previous hypothesis that the relative distances between predicate and its arguments are within a certain range. Besides, $(0,1)$ is the most frequently occurring relationships in all languages except Japanese, which indicates that arguments most frequently appear to be the children of their predicate in dependency syntax tree. 

Figure \ref{fig:topk} shows F$_1$ scores on test set  by top-$k$ argument pruning for German, Catalan and Japanese,\footnote{Note that the maximum values of $k$ for German, Catalan and Japanese are 36, 50 and 260, respectively. We thus show only the top 50 for Japanese due to the limited space.} where $k$ = 0 represents the baseline without pruning. We observe that the case of $k$ = 20 yields the best performance on German and Catalan. As for Japanese, it falls short of the baseline in our current observation range, but our experiment has shown that the setting of top-120 can achieve the best results.

\begin{table}
	\centering
	\begin{tabular}{lccc}
		\toprule
		Dataset & syntactic rule & $k$-order & $\Delta$ F$_1$\\
		\midrule
		Catalan & 84.35 & 84.18 & $-$0.17 \\
		Czech & 88.76 & 88.75 & $-$0.01 \\
		German & 78.54 & 78.42 & $-$0.12\\
		Japanese & 83.12 & 82.61 & $-$0.51 \\
		Spanish & 83.70 & 83.53 & $-$0.17\\
		\bottomrule
	\end{tabular}
	\caption{Comparison of our model with syntactic rule and $k$-order argument pruning.}\label{tab:k-order}
\end{table}

\begin{table}
	\centering
	\setlength{\tabcolsep}{5pt}
	\begin{tabular}{lccc}
		\toprule
		\multirow{2}{*}{Dataset}&\multirow{2}{*}{syntax-}&\multicolumn{2}{c}{syntax-aware}\\ 
		\cmidrule(lr){3-4}
		&agnostic&predicted \small{(UAS)}&\textit{gold}\\
		\midrule
		Catalan & 84.07 & 84.35 \small{(89.43)} & \textit{85.50} \\
		Czech & 88.35 & 88.76 \small{(85.69)} & \textit{88.92} \\
		German & 78.36 & 78.54 \small{(88.91)} & \textit{78.56} \\
		Japanese & 83.08 & 83.12 \small{(92.29)} & \textit{83.20} \\
		Spanish & 83.47 & 83.70 \small{(89.39)} & \textit{84.82} \\
		\bottomrule
	\end{tabular}
	\caption{Syntactic contribution to multilingual SRL. \textit{predicted} and \textit{gold} denote the use of syntactic parse. The UAS of predicted syntax is in parenthesis.}\label{tab:gold}
\end{table}

To reveal the strengths of proposed syntactic rule, we conduct further experiments, replacing our syntactic rule based pruning with the $k$-order argument pruning of \citet{he:2018Syntax}. Following their setting, we use the tenth-order pruning for pursuing the best performance. Table \ref{tab:k-order} shows the performance gaps between two pruning methods. Comparing with syntactic rule, the $k$-order pruning declines model performance by 0.2\% F$_1$ on average, showing that our syntactic rule based pruning method is more effective and can be extended well to multiple languages especially for Japanese.

\subsection{Syntactic Impact}
In this part, we attempt to explore the syntactic impact on other five languages. To investigate the most contribution of syntax to multilingual SRL, we perform experiments using the gold syntactic parse also officially provided by the CoNLL-2009 shared task instead of the predicted one.\footnote{In this work, we use gold syntax rather than other better parse to explore the greatest syntactic contribution, considering the current state-of-the-art syntactic parsers are being upgraded so fast now. } To be more precise, the syntactic rule is counted based on gold syntactic tree and applied to argument pruning layer. The corresponding results of our syntax-agnostic and syntax-aware models are summarized in Table \ref{tab:gold}. We also report the unlabeled attachment scores (UAS) of predicted syntax as syntactic accuracy measurement, considering that we do not use the dependency labels.

Results indicate that high-quality syntax can further improve model performance, showing syntactic information is generally effective for multilingual SRL. In particular, based on gold syntax, the top-$1$ argument pruning for Catalan and Spanish has reached 100 percent coverage (namely, for Catalan and Spanish, all arguments are the children of predicates in gold dependency syntax tree), and hence our syntax-aware model obtains significant gains of 1.43\% and 1.35\%, respectively. In addition, combining the results of Tables \ref{tab:k-order} and \ref{tab:gold}, we find that applying the $k$-order pruning to syntax-agnostic model results in better performance on most languages. However, \citet{cai2018full} argue that $k$-order pruning does not boost the performance for English. One reason to account for this finding is the lack of effective approaches for incorporating syntactic information into sequential neural networks. Nevertheless, syntactic contribution is overall limited for multilingual SRL in this work, due to strong syntax-agnostic baseline. Therefore, more effective methods to incorporate syntax into neural SRL model are worth exploring and we leave it for future work.

\section{Related Work}
In early work of semantic role labeling, most of researchers were dedicated to feature engineering \cite{pradhan2005,punyakanok2008importance,zhao2009,zhao-jair-2013}. The first neural SRL model was proposed by \citet{Collobert2011}, which used convolutional neural network but their efforts fell short. Later, \citet{Foland2015} effectively extended their work by using syntactic features as input. \citet{roth2016} introduced syntactic paths to guide neural architectures for dependency SRL. 

However, putting syntax aside has sparked much research interest since \citet{zhou-xu2015} employed deep BiLSTMs for span SRL. A series of neural SRL models without syntactic inputs were proposed. \citet{marcheggiani2017} applied a simple LSTM model with effective word representation, achieving encouraging results on English, Chinese, Czech and Spanish. \citet{cai2018full} built a full end-to-end SRL model with biaffine attention and provided strong performance on English and Chinese. \citet{lizc2019srl} also proposed an end-to-end model for both dependency and span SRL with a unified argument representation, obtaining favorable results on English.

Despite the success of syntax-agnostic SRL models, more recent work attempts to further improve performance by integrating syntactic information, with the impressive success of deep neural networks in dependency parsing \cite{zhang2016-probabilistic,zhou-2019-head}. \citet{marcheggianiEMNLP2017} used graph convolutional network to encode syntax into dependency SRL. \citet{he:2018Syntax} proposed an extended $k$-order argument pruning algorithm based on syntactic tree and boosted SRL performance. \citet{li2018emnlp} presented a unified neural framework to provide multiple methods for syntactic integration. Our method is closely related to the one of \citet{he:2018Syntax}, designed to prune as many unlikely arguments as possible.

\paragraph{Multilingual SRL}
To promote NLP applications, the CoNLL-2009 shared task advocated performing SRL for multiple languages. Among the participating systems, \citet{Zhao2009Conll} proposed an integrated approach by exploiting large-scale feature set, while \citet{bjorkelund2009} used a generic feature selection procedure. Until now, only a few of work \cite{Lei2015,swayamdipta2016,polyglot2018acl} seriously considered multilingual SRL. Among them, \citet{polyglot2018acl} built a polyglot model (training one model on multiple languages) for multilingual SRL, but their results were far from satisfactory. Therefore, this work aims to complete the overall upgrade since CoNLL-2009 shared task and leaves polyglot training as our future work.

\section{Conclusion}
This paper is dedicated to filling the long-term performance gap of multilingual SRL since a long time ago with a newly proposed syntax-based argument pruning method. Experimental results demonstrate its effectiveness and shed light on a new perspective for many NLP tasks to incorporate syntax simply and effectively. Besides, our model substantially boosts multilingual SRL performance by introducing deep enhanced representation, achieving new state-of-the-art results on the in-domain CoNLL-2009 benchmark for Catalan, Chinese, Czech, English, German, Japanese and Spanish. These results further show that syntactic information and deep enhanced representation can also promote multiple languages rather than only the case of English.

\bibliography{emnlp-ijcnlp-2019}
\bibliographystyle{acl_natbib}

\end{document}